\pdfoutput=1
\documentclass[11pt]{article}

\usepackage{ACL2023}

\usepackage{times}
\usepackage{latexsym}
\usepackage{subcaption}
\usepackage{graphicx}
\usepackage{multirow}
\usepackage{multicol}
\usepackage{booktabs}

\usepackage[T1]{fontenc}
\usepackage[utf8]{inputenc}

\usepackage{microtype}

\usepackage{inconsolata}

\title{PreCog: Exploring the Relation between Memorization\\ and Performance in Pre-trained Language Models}

\author{
  Leonardo Ranaldi ($\bullet$), 
   \textbf{Elena Sofia Ruzzetti}, and  
  \textbf{Fabio Massimo Zanzotto}\\ ART Group Department of Enterprise Engineering,\\
  University of Rome Tor Vergata,\\
  Viale del Politecnico, 1, 00133 Rome, Italy;\\
  ($\bullet$)\texttt{\{leonardo.ranaldi\}@uniroma2.it} \\
}

\begin{document}

\maketitle
\begin{abstract}
Pre-trained language models such as BERT are impressive machines with the ability to memorize, possibly generalized learning examples. We present here a small, focused contribution to the analysis of the interplay between memorization and performance of BERT in downstream tasks. We propose \emph{PreCog}, a measure for evaluating memorization from pre-training, and we analyze its correlation with the BERT's performance.  Our experiments show that highly memorized examples are better classified, suggesting memorization is an essential key to success for BERT.

\end{abstract}

\section{Introduction}

%\mainideafortheparagraph{pre-training these models on large corpora yields great lexical knowledge}

Pre-trained language models (PTLMs) \citep{peters-etal-2018-deep,devlin-etal-2019-bert,Liu2019RoBERTaAR} are intriguing machines dominating the arena of NLP tasks with their ability to memorize generalizations of texts in synthetic neurons. After long pre-training on large amounts of unlabeled data, PTLMs have shown to learn effectively downstream tasks with limited labeled data  \citep{Howard2018UniversalLM} and generalize in out-of-distribution examples \citep{hendrycks-etal-2020-pretrained}. Extensive studies have shown that these PTLMs tend to mimic traditional linguistic syntactic models \citep{jawahar-etal-2019-bert} and traditional NLP pipelines \citep{tenney-etal-2019-bert}. Hence, a crucial issue is to clarify why PLTMs exploit pre-training better than traditional NLP modules exploit annotated corpora.

%Recent advances in pre-trained language models \citep{peters-etal-2018-deep,devlin-etal-2019-bert,Liu2019RoBERTaAR}, the field of computational linguistics has seen improvements in a wide range of tasks and applications.
%Downstream of long pre-training phases, these models have gained general knowledge from large amounts of unlabeled data. Moreover, such methods have been shown to learn effectively with limited labeled data for downstream tasks \citep{Howard2018UniversalLM} and generalize in out-of-distribution examples \citep{hendrycks-etal-2020-pretrained}. Previous work has extensively studied linguistic \citep{jawahar-etal-2019-bert} and relational \citep{tenney-etal-2019-bert} types of knowledge. 
%However, the process of learning these models from downstream data and the qualitative nature of their learning dynamics still need to be clarified. 

%\mainideafortheparagraph{pre-training produces general knowledge. instead, it seems to induce memorization}

Understanding the learning process of PTLMs may help in understanding their results in downstream tasks and in improving their linguistic representations in scenarios where they fail \cite{Kumar2020UserGD}. Indeed, unlike traditional general NLP modules in pipelines, PTLMs need to be fine-tuned for the specific tasks \citep{devlin-etal-2019-bert} and, eventually, domain-adapted on the specific language of the novel corpus \cite{jin-etal-2022-lifelong-pretraining}. Moreover, as many other machine learning models, fine-tuned PTLMs lose their ability to solve a task if subsequently fine-tuned to another task \cite{CatastrophicForhetting:BERT:9206891} although they apparently do not change their language models \cite{merchant-etal-2020-happens}. This phenomenon is known as \emph{catastrophic forgetting} \cite{Catastrophic:doi:10.1073/pnas.1611835114} in machine learning. Then, it is still unclear how these models exploit pre-training and training examples.

%{Understanding of the learning process of PTLMs may help in understanding their results in downstream tasks}{and may help in understandandin in which scenarios they will fail and how to improve them to achieve more robust linguistic representations.}{Indeed, ... The process of fine-tuning Pre-trained Language Models (PLMs) such as BERT \citep{devlin-etal-2019-bert} aims to challenge the mnemonic capabilities of pre-training by adapting the model to the general case.  
%For many applications, the model needs to be able to generalize and remember only some things by rote. PLMs should generalize the knowledge learned during the pre-training phase for other applications.
%In some cases, it is helpful for the model to be generalized, so that common patterns of the task are learned while discarding irrelevant noise and outliers. 
%On the other side of the coin, discarding everything that occurs infrequently is not a trustworthy learning strategy. In many low-resource scenarios, memorization may be the key to good results in a downstream task \citep{tu-etal-2020-empirical}.

PTLMs, such as BERT \citep{devlin-etal-2019-bert}, have shown to have an impressive ability to memorize and possibly generalize learning examples. This ability has been largely investigated as it may be extremely harmful. In fact, these PTLMs may reveal sensitive information that has been acquired during pre-training. For example, memories of Generative Pretrained Transformers (GPTs) \citep{Radford2018ImprovingLU-GPT-Base} have been violated and produced phone numbers, and usernames \citep{DBLP:conf/uss/CarliniTWJHLRBS21,Thakkar-UnderstandingUnintended-gpt}. However, this simple ability to memorize may play a crucial role in the performances of PTLMs in downstream tasks.

This paper presents a small, focused contribution to the role of memorization in the performance of BERT in downstream tasks. We propose \emph{PreCog}, a very simple measure of coverage that evaluates how much pre-training covers the information needed to model a given example or, better if BERT has already partially seen the example - it \emph{pre}-cognizes the example. The aim is to evaluate if PreCog precognizes on which examples BERT adapted to a downstream task performs better inferences. We have extensively experimented with PreCog by using BERT over the GLUE tasks \cite{wang-etal-2018-glue}, and we observed the ability of PreCog to predict examples where a task-adapted BERT performs better. Besides being a predictive measure, PreCog showed that example memorization is a crucial part of the success of BERT. 

%By constructing experiments that study sentences seen in pre-training, we can explore the dynamics of pattern learning under conditions of memorization and generalization of PLMs.
%To our knowledge, this is the first qualitative study of the behavior of transformer-based PLMs under conditions of varying input sentences based on knowledge gained during the pre-training phase. 

%\mainideafortheparagraph{Our Proposal: Methodology for analyzing whether sentences were seen in the pre-training phase}
%Our contributions are the following: 1)We propose a method to estimate the coverage of a sentence or, better if it has already been seen in the pre-training phase. 2)We then apply the coverage function to estimate the average coverage of a set of downstream tasks. 3) We find that models such as BERT are particularly good at learning even in the absence of memorized knowledge during pre-training. 4) We empirically observe that despite the fine-tuning phase, a distinct performance plateau occurs for different values of knowledge acquired during pre-training. 
%Finally, we show how as the average coverage decreases, the average accuracies decrease, showing how much pre-training affects the performances of downstream tasks.

\newpage

\section{Related Work}

The ability of linguistic neural models to memorize facts is out of doubt. This ability has been deeply explored as it is a problem for privacy issues. Indeed, LSTM language models remember facts so well that individual facts can be retrieved during inference \cite{Carlini2019TheSS}. These facts may reveal sensitive personal information such as names and addresses associated with people. Moreover, revitalizing the idea of sparse distributed memories \cite{kanerva1988sparsedistributedmemory}, \citet{Petroni2019LanguageMA} hypothesized that large language models might be used as clever and inexpensive ways to build up effortlessly knowledge bases. Even in other areas like image classification, it appears that large neural networks may memorize entire datasets as these networks achieve very low error rates over datasets with random generated target labels \cite{DBLP:conf/iclr/ZhangBHRV17}. Yet, it is still unclear to what extent this ability to memorize facts helps neural networks in downstream tasks.
%Research on the ability of linguistic neural models to memorize and remember facts seen during their training is ample.
%\citet{Carlini2019TheSS}, studying LSTM language models, showed that they can store individual out-of-distribution examples during the training and that it is possible to recover these examples at the moment of testing.

%\citet{Forgetting}, on the other hand, study forgetting process occurs as a model learns examples finding that models consistently forget a significant fraction of training data. Furthermore, the portion of training data that is not remembered depends on the intrinsic properties of the training data rather than on the specific model.

%At the same time, \citet{Petroni2019LanguageMA} showed that pre-trained language models surprisingly effectively recall facts.

A key research question is to understand how large pre-trained neural networks generalize over memorized examples. Pre-training seems to be a winning strategy to boost generalization. In fact, pre-trained models generalize better on out-of-distribution data and can detect such data better than non-pre-trained methods \cite{hendrycks-etal-2020-pretrained}. However, these models need a significant number of training instances to exploit this generalization ability in downstream tasks \cite{tanzer-etal-2022-memorisation}. Hence, since fine-tuning on specific datasets seems to be connected to  \textit{catastrophically forgetting} examples \cite{CatastrophicForhetting:BERT:9206891}, generalization and memorization can be strictly correlated.
%state that the number of examples significantly affects the learning process, influencing both the time at which examples are stored and the quality of generalization.
%However, despite the current theoretical knowledge of statistical learning, it is challenging to clarify the superhuman generalization performance of large neural models \citep{10.1145/3446776}.

%However, they still need to cleanly separate in- and out-of-distribution examples.

%The effective capacity of neural networks is sufficient for memorizing the entire data set
%Memorization is nearly interconnected to generalization: it has been observed that neural networks learn simple patterns before noise and generalize despite being able to memorize random examples \citep{DBLP:conf/iclr/ZhangBHRV17}.

% Nevertheless, at the same time, pre-trained methods such as BERT are sensitive to spelling noise and typos \citep{Kumar2020UserGD}. The generalization effect could be a hook effect, as remarked in \citep{heinzerling2019cleverhans}. The hook effect would affect the storage part by blurring the generalization, especially when there is a lot of data available. 

%We define a “forgetting event” to have occurred when an individual training example transitions from being classified correctly to incorrectly over the course of learning. \citet{Forgetting}

{To explore the correlation between memorization and performance on downstream tasks,}{we propose a mechanism for analyzing sentence coverage.}{In particular, we investigate how much sentences are seen in the pre-training phase in transformer-based PLMs using perturbation masking methods. These methods allow us to observe the impact of pre-training on the performance of downstream tasks.}{This novel measure is needed as current measures for understanding coverage, such as  “forgetting event” \cite{Forgetting} and counterfactual memorization \cite{Zhang2021CounterfactualMI}, mix performance and actual memorization.}

\section{Method and Data}

This section introduces PreCog that is our measure to evaluate how much pre-training covers the information needed to model a given example (Sec. \ref{sec:precog}), two comparative measures $Lenght$ and $LexCov$ (Sec. \ref{sec:compar}), and the experimental setting  (Sec. \ref{sec:tasks}).

\begin{figure*}
\centering
\begin{subfigure}{0.3\textwidth}
    \includegraphics[width=\textwidth]{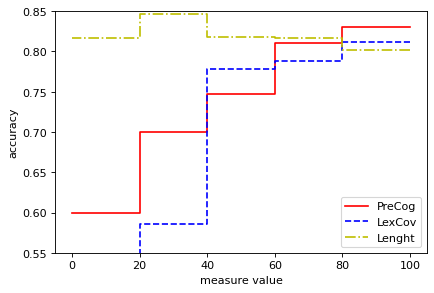}
    \caption{Accuracy histograms of $BERT_{FT}$ on bins of 20 points plotted vs. the value of measures.}
    \label{fig:bins}
\end{subfigure}
\hfill
\begin{subfigure}{0.3\textwidth}
    \includegraphics[width=\textwidth]{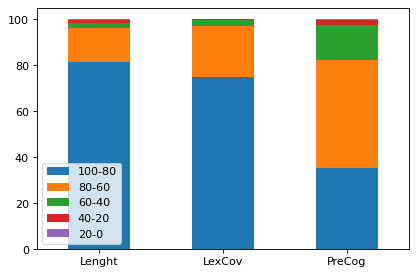}
    \caption{Percent of coverage of the dataset for intervals of values of the measures. }
    \label{fig:perc_cover}
\end{subfigure}
\hfill
\begin{subfigure}{0.3\textwidth}
    \includegraphics[width=\textwidth]{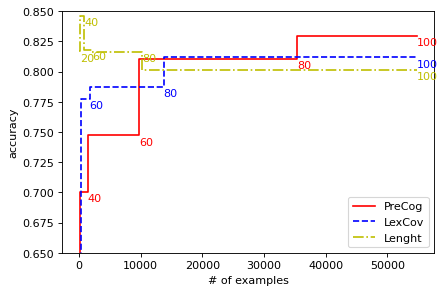}
    \caption{Accuracy histograms of $BERT_{FT}$ on bins of 20 points plotted vs. the coverage of the testset of the bins for the three measures. Values of the extremes of the bins are reported on the curve related to the measure.}
    \label{fig:coverage}
\end{subfigure}
\caption{Accuracy plots of $BERT_{FT}$ for the weighted sum of accuracies in each GLUE task.}
\label{fig:figures}
\end{figure*}

\subsection{\emph{PreCog}: a measure to evaluate pre-training coverage}
\label{sec:precog}
%We propose the perturbed masking technique to assess the MLM task's impact during pre-training. The information between the derived words is the basis for our subsequent analysis.

%\subsubsection{MLM in Action}

%\subsubsection{\emph{PreCog}: a measure to evaluate pre-training coverage}

{BERT \cite{devlin-etal-2019-bert} is pre-trained on billions of text tokens}{by using the Masked Language Modeling (MLM) as one of the two main learning tasks.}{Indeed, during pre-training, MLM randomly selects and masks 15\% of all tokens in any given sequence. This 15\% of tokens are either (a) replaced with the special token [MASK], (b) replaced by a random token, or (c) kept unchanged with a respective probability of 80\%, 10\%, and 10\%.  Then, BERT learns to predict the masked tokens. 
%During the masking process, a token is either (a) replaced with the special token [MASK], (b) replaced by a random token, or (c) kept unchanged. These phenomena occur at a random rate of 80\%, 10\%, and 10\%.
This task is learned till near the overfitting.}{Then, one of the main ability of BERT is unmasking masked tokens.}

{We aim to capture}{to which extent a sequence of tokens is covered by pre-training in transformers such as BERT .}{For this reason, we build on the core capacity of BERT, that is, unmasking masked tokens. Hence, if BERT can predict masked tokens of a given sequence of tokens, it possibly has the knowledge to better deal with that sequence.}{Our intuition is that a measure built on  unmasking masked tokens describes the ``prior'' knowledge of BERT over sequences. }

%(2) Prediction of the next sentence: given a pair of sentences predict whether the second sentence follows the first sentence in an original document or is taken from another random document.

{Given a sentence or text excerpt as a list of tokens $x = [x_1, . . . , x_T]$, our function $PreCog(x)$ is defined as follows.}{Firstly, we mask one by one each token in $x$ obtaining T different sequences $\hat{x}_i = [x_1, ..., x_{i-1}, [MASK], x_{i+1}. . , x_T]$.  Then, the measure is straightforwardly defined as:\\
\begin{equation}
    PreCog_l(x) = \frac{\sum_{i=0}^{T} \delta(x_i\in BERT_{MLM}(\hat{x}_i))}{T}  
\end{equation}
where $BERT_{MLM}(\hat{x}_i)$ is the set of the first $100$ tokens predicted by BERT for the  position $i$ and $\delta(x_i \in X)$ is 1 if $x_i \in X$ and 0 otherwise.}

%Given a sentence or text excerpt as a list of tokens $x = [x_1, . . . , x_T]$, BERT maps each $x_i$ into a contextualized representation $H_{\theta}(x)_i$, where $\theta$ represents the network parameters. We aim to derive a function $f(x)$ that captures the impact of individual tokens $x_i$ on the contextualized word $x$.
%For each sentence, we propose a three-stage approach to achieve our goal. First, sequentially we replace $x_i$ with the token [MASK]. Second, given the masked judgments we ask BERT to guess the tokens that have been masked. If the masked token is guessed by BERT $1$, otherwise $0$. Finally, we define the coverage function $f(x)$ as: 
%\[ 
%\]  

{PreCog is a very simple measure.}{Yet, it may reveal important facts about how BERT uses pre-training text in downstream tasks.}{A very important issue is to understand if PreCog correlates with the performance of BERT in these tasks.}{A positive and steady correlation will be an important hint for understanding the role of pre-training.}

\subsection{Alternative Coverage Measures}
\label{sec:compar}

To comparatively evaluate $PreCog$, we use two measures: Length and LexCov. Length aims to correlate the accuracy of BERT to the length of samples and LexCov to the coverage of dictionary of BERT. Then, the measures are defined as follows:
\begin{itemize}
    \item $Length(x) = \frac{T - min_D}{max_D - min_D}$ where T is the length of $x$, $min_D$ and $max_D$ are the min and the max length of samples in a dataset $D$; 
    \item $LexCov(x) = \frac{T - |OOV(x)|}{T}$ where $OOV(x)$ is the set of the out-of-vocabulary words of the example $x$ with respect to BERT's vocabulary.
\end{itemize}

%\subsubsection{Token Perturbation on downstream tasks}

%In order to analyze the performance of pre-trained models on different values of sentence coverage, we used the famous General Language Understanding Evaluation (GLUE) dataset \cite{wang-etal-2018-glue}. For each dataset, we estimated the coverage of test-sets.
%This choice was made both to have terms of comparison with the state of the art and for ease of retrieval in the huggingface library \cite{hugging}. 

%\section{Experimental Setup}

\subsection{Experimental set-up}
\label{sec:tasks}

To experiment with a variety of tasks, we use the GLUE benchmark \citep{wang-etal-2018-glue} containing tasks for: (1) natural language inference, that is,  Multigenre NLI (MNLI) \citep{williams-etal-2018-broad}, Question NLI (QNLI) \citep{wang-etal-2018-glue}, Recognizing Textual Entailment (RTE) \citep{DBLP:conf/tac/BentivogliMDDG09}, and Winograd NLI (WNLI) \citep{10.5555/3031843.3031909}; (2)  semantic similarity, that is, the Microsoft Research Paraphrase Corpus (MRPC) \citep{dolan-brockett-2005-automatically}, the Semantic Textual Similarity Benchmark (STS-B) \citep{cer-etal-2017-semeval}, and Quora Question Pairs (QQP) \citep{Sharma2019NaturalLU}; sentiment classification - Stanford Sentiment Treebank (SST-2) \citep{socher-etal-2013-recursive}; and corpus of linguistic acceptability (CoLA) \citep{warstadt-etal-2019-neural}. 
%Precisely, NLI is covered by Multigenre NLI (MNLI) \citep{williams-etal-2018-broad}, Question NLI (QNLI) \citep{wang-etal-2018-glue}, Recognizing Textual Entailment (RTE) \citep{DBLP:conf/tac/BentivogliMDDG09}, and Winograd NLI (WNLI) \citep{10.5555/3031843.3031909}. Semantic similarity is represented by the Microsoft Research Paraphrase Corpus (MRPC) \citep{dolan-brockett-2005-automatically}, the Semantic Textual Similarity Benchmark (STS-B) \citep{cer-etal-2017-semeval}, and Quora Question Pairs (QQP) \citep{Sharma2019NaturalLU}. Finally, sentiment classification consists of the Stanford Sentiment Treebank (SST-2) \citep{socher-etal-2013-recursive} and linguistic acceptability is covered by CoLA \citep{warstadt-etal-2019-neural}. 
SST-2 and CoLA are single sentence tasks.

%To experiment with a variety of different tasks, we use the GLUE benchmark \citep{wang-etal-2018-glue} containing tasks for natural language inference (NLI),  semantic similarity, sentiment classification, and linguistic acceptability. 
%Precisely, NLI is covered by Multigenre NLI (MNLI) \citep{williams-etal-2018-broad}, Question NLI (QNLI) \citep{wang-etal-2018-glue}, Recognizing Textual Entailment (RTE) \citep{DBLP:conf/tac/BentivogliMDDG09}, and Winograd NLI (WNLI) \citep{10.5555/3031843.3031909}. Semantic similarity is represented by the Microsoft Research Paraphrase Corpus (MRPC) \citep{dolan-brockett-2005-automatically}, the Semantic Textual Similarity Benchmark (STS-B) \citep{cer-etal-2017-semeval}, and Quora Question Pairs (QQP) \citep{Sharma2019NaturalLU}. Finally, sentiment classification consists of the Stanford Sentiment Treebank (SST-2) \citep{socher-etal-2013-recursive} and linguistic acceptability is covered by CoLA \citep{warstadt-etal-2019-neural}. SST-2 and CoLA are single sentence tasks. 

%\subsection{Selected BERT and Test Procedure}
%\label{sec:ptlm}

We used two version of BERT \citep{devlin-etal-2019-bert}: $BERT_{FT}$ with fine-tuning and $BERT_{DA}$ with domain-adaptation. These two are based on the pre-trained  version of BERTforSequenceClassification (see \cite{wolf-etal-2020-transformers}). 
%XLNet \citep{NEURIPS2019_dc6a7e65} is an alternative; however, it uses a slightly different attention structure and masked language modeling algorithm. Our preliminary experiments encountered reproducibility difficulties with the Transformers library.
%Each model has a basic and a large variant containing 12 and 24 layers, respectively. We denote these by adding the variant name as a subscript to the model name.
The fine-tuning procedure is that of traditional BERT. For each downstream task, we chose the Adam optimizer \cite{Kingma2015AdamAM} with a batch size of $16$ and fine-tuned BERT for 4 epochs, following the original paper \cite{devlin-etal-2019-bert}. For hyperparameter tuning, the best learning rate is different for each task, and all original authors choose one between $1\times10^{-5}$ and $5\times10^{-5}$.

We conduct our experiments on NVIDIA RTX A6000 GPUs with CUDA v11.3. We run the models from the Transformers library \cite{wolf-etal-2020-transformers} using PyTorch v1.12.0.

To study the correlation between the performance of BERT on the one side and one of the three measures - PreCog, Length, or LexCov - on the other side, we divided the sequences $x$ in testsets in 5 bins according to the value of the measure, we plotted histograms of accuracies of BERT with respect to the three measures (Fig. \ref{fig:figures}), and we computed the Pearson's correlation of the measure with respect to the accuracies (Tab. \ref{tab:correlations}).

%Our testing procedure is based on the coverage values defined in Section X. to produce comparable results. We collected coverage values in 5 rounds with equidistant intervals. 
%In addition, we reported the performance on the entire test dataset.

\begingroup
\tabcolsep = 4.0pt
\def\arraystretch{0.50}

\begin{table*}[]
    \centering
    \tiny

\begin{tabular}{lcc|c|rrr|rrr|rrr}
\toprule
      & \multicolumn{2}{c}{Global}  &    &         \multicolumn{3}{c|}{Length} &         \multicolumn{3}{c|}{LexCov} & \multicolumn{3}{c}{PreCog} \\
 Task & $BERT_{FT}$ &  $BERT_{DA}$ &     interval &  \# samples &  $BERT_{FT}$ &  $BERT_{DA}$ &     \# samples &  $BERT_{FT}$ &  $BERT_{DA}$ &  \# samples &  $BERT_{FT}$ &  $BERT_{DA}$ \\
\midrule
 \multirow{2}*{CoLa} & \multirow{2}*{0.920} & \multirow{2}*{0.935} &  (80,100] &    499 &    0.906 & 0.918 &    857 & 0.926 & 0.940 &    577 & 0.951 & 0.972 \\
      &&&    [0,80] &    446 &    0.935 & 0.955 &     88 & 0.852 & 0.886 &    368 & 0.870 & 0.878 \\
%\cmidrule(lr){2-11}
%      &&&   0-100 &    945 &    0.920 & 0.935 &    945 & 0.920 & 0.935 &    945 & 0.920 & 0.935 \\
\midrule
 \multirow{2}*{mnli} & \multirow{2}*{0.716} & \multirow{2}*{0.721} &  (80,100] &   7782 &    0.717 & 0.721 &   6512 & 0.739 & 0.745 &   3508 & 0.759 & 0.770 \\
      &&&    [0,80] &   1361 &    0.716 & 0.718 &   2631 & 0.660 & 0.660 &   5635 & 0.690 & 0.690 \\
%\cmidrule(lr){2-11}
%      &&&   0-100 &   9143 &    0.716 & 0.721 &   9143 & 0.716 & 0.721 &   9143 & 0.716 & 0.721 \\
\midrule
 \multirow{2}*{mrpc} & \multirow{2}*{0.806} & \multirow{2}*{0.861}&  (80,100] &     59 &    0.780 & 0.831 &    924 & 0.818 & 0.877 &    376 & 0.867 & 0.880 \\
      &&&    [0,80] &   1590 &    0.806 & 0.861 &    725 & 0.789 & 0.839 &   1273 & 0.787 & 0.854 \\
%\cmidrule(lr){2-11}
%      &&&   0-100 &   1649 &    0.805 & 0.860 &   1649 & 0.805 & 0.860 &   1649 & 0.805 & 0.860 \\
\midrule
 \multirow{2}*{qnli} &\multirow{2}*{0.808} & \multirow{2}*{0.829}&  (80,100] &   3245 &    0.802 & 0.832 &   3123 & 0.809 & 0.831 &   1769 & 0.832 & 0.846 \\
      &&&    [0,80] &   1970 &    0.817 & 0.825 &   2092 & 0.807 & 0.827 &   3446 & 0.796 & 0.821 \\
%\cmidrule(lr){2-11}
%      &&&   0-100 &   5215 &    0.808 & 0.829 &   5215 & 0.808 & 0.829 &   5215 & 0.808 & 0.829 \\
\midrule
  \multirow{2}*{qqp} &\multirow{2}*{0.822} & \multirow{2}*{0.845}&  (80,100] &  32728 &    0.820 & 0.845 &  28862 & 0.823 & 0.843 &  12810 & 0.840 & 0.860 \\
      &&&    [0,80] &   3990 &    0.834 & 0.842 &   7856 & 0.816 & 0.850 &  23908 & 0.812 & 0.837 \\
%\cmidrule(lr){2-11}
%      &&&   0-100 &  36718 &    0.822 & 0.845 &  36718 & 0.822 & 0.845 &  36718 & 0.822 & 0.845 \\
\midrule
  \multirow{2}*{rte} &\multirow{2}*{0.646} & \multirow{2}*{0.653}&  (80,100] &    146 &    0.671 & 0.678 &    155 & 0.716 & 0.723 &     46 & 0.652 & 0.674 \\
      &&&    [0,80] &    122 &    0.615 & 0.623 &    113 & 0.549 & 0.558 &    222 & 0.644 & 0.649 \\
%\cmidrule(lr){2-11}
%      &&&   0-100 &    268 &    0.646 & 0.653 &    268 & 0.646 & 0.653 &    268 & 0.646 & 0.653 \\
\midrule
 \multirow{2}*{sst2} &\multirow{2}*{0.939} & \multirow{2}*{0.924}&  (80,100] &    151 &    0.907 & 0.887 &    607 & 0.951 & 0.946 &    333 & 0.970 & 0.970 \\
      &&&    [0,80] &    655 &    0.947 & 0.933 &    199 & 0.905 & 0.859 &    473 & 0.918 & 0.892 \\
%\cmidrule(lr){2-11}
%      &&&   0-100 &    806 &    0.939 & 0.924 &    806 & 0.939 & 0.924 &    806 & 0.939 & 0.924 \\
\midrule
 \multirow{2}*{wnli} & \multirow{2}*{0.565} & \multirow{2}*{0.594}&  (80,100] &     31 &    0.452 & 0.484 &     61 & 0.590 & 0.623 &     39 & 0.590 & 0.615 \\
      &&&    [0,80] &     38 &    0.658 & 0.684 &      8 & 0.375 & 0.375 &     30 & 0.533 & 0.567 \\
%\cmidrule(lr){2-11}
%      &&&   0-100 &     69 &    0.565 & 0.594 &     69 & 0.565 & 0.594 &     69 & 0.565 & 0.594 \\
\bottomrule
\end{tabular}

    \caption{Accuracies on the GLUE tasks computed grouping datasets according to the values of three measueres - PreCog, LexCov, and Lenght - for $BERT_{FT}$ and $BERT_{DA}$.}
    \label{tab:task_results}
\end{table*}

\section{Experimental Results and Discussion}

Accuracies reported in Fig. \ref{fig:bins} and Fig. \ref{fig:coverage} and used in Tab. \ref{tab:correlations} are the weighted sum of accuracies in each GLUE task.
%of evaluated on the combination of the testing sets of tasks in GLUE whereas the training of $BERT_{FT}$ is performed separately on each task. 
This guarantees that the 20-point bins have a sufficient set of samples to compute stable accuracies. 

\begin{table}[h!]
    \centering
    \begin{small}
    
    \begin{tabular}{l|cc}
    \toprule
       \emph{Measure}  & \emph{Correlation} & \emph{p-value} \\
\midrule
Length & -0.5922 & 0.292 \\
LexCov & 0.9014 & 0.037\\
PreCog &  0.9737 & 0.005\\
\bottomrule
    \end{tabular}
    \end{small}
    \caption{Pearson's correlation between the measures and the accuracy bins of $BERT_{FT}$ for the combined GLUE tasks. }
    \label{tab:correlations}
    \vspace{-5mm}
%(0.9737523474134151, 0.0050845469378595605), (0.7886371375338975, 0.11287680728246317), (0.7884869035108621, 0.11299443831661167), (0.8740257714211479, 0.05264715792420113), (0.9013838014739027, 0.03662059825643528), (-0.5922059888162992, 0.29272470366806397)
\end{table}

PreCog correlates with the accuracy of $BERT_{FT}$ better than Lenght and LexCov (see Fig. \ref{fig:bins} and Tab. \ref{tab:correlations}). Accuracies of PreCog in the different bins degrade more uniformly than the other two measures (red solid line in Fig. \ref{fig:bins}). Moreover, the Pearson's correlation between PreCog values and the accuracies of $BERT_{FT}$ is 0.9737 with a p-value of 0.005 and it is higher than the ones of both LexCov, 0.9014 with a p-value of 0.037, and Length which is not correlated (see Tab. \ref{tab:correlations}).

PreCog values better separate examples in testing sets. At first glance, LexCov may seem a better model to separate samples with high with respect to those with less accuracy expectations. Samples with a value of LexCov less than 40 have low accuracy (see Fig. \ref{fig:bins}). However, samples having LexCov between 0 and 40 are rare (Fig. \ref{fig:perc_cover}). Better observations are derived by plotting accuracies over bins rescaled according to their coverage (Fig. \ref{fig:coverage}). Indeed, PreCog separates samples better than LexCov (red solid line vs. dashed blue line in Fig. \ref{fig:coverage}): samples from 18,000 to 55,000 fall in two bins for PreCog and in only one bin for LexCov. Hence, PreCog has better discriminative power than LexCov.

Results are substantially confirmed on task basis: PreCog is a better predictor of the accuracy on tasks and a better separator of classes of samples (see Tab. \ref{tab:task_results}). Accuracies of $BERT_{FT}$ are generally higher for samples with PreCog in the interval $[80,100]$ than for samples with the other two measures in the same interval. $LexCov$ has higher accuracy for samples in $[80,100]$ only for RTE. Moreover, accuracies of samples in the interval $[80,100]$ are always higher than those in the interval $[0,80]$ for both PreCog and LexCov. Yet, PreCog partitions more evenly samples and the differences in accuracies between intervals $[80,100]$ and $[0,80]$ are generally higher.     

Moreover, domain adaptation is not changing the above findings. Accuracies for $BERT_{DA}$ are generally higher than those without domain adaptation for all the tasks except for SST2 and WNLI (Tab. \ref{tab:correlations}). Moreover, focusing on PreCog, the overall increase in accuracies in CoLa, MNLI, and RTE derives from an increase in the samples of the interval $[80,100]$. This fact suggests that $BERT_{DA}$ is gaining a better model for these samples.    

As a final observation, BERT seems to behave better on sentences that have been, at least, partially seen during pre-training. Indeed, PreCog is a measure capturing how much the sentence is covered with the pre-training task Masked Language Model (MLM). Typically, BERT overfits on MLM during pre-training. Then, PreCog is a measure telling whether sentences have already been partially seen. Instead, LexCov describes how many words of sentences are covered by BERT's vocabulary. Since there is a great difference in predicting accuracy on tasks between PreCog and LexCov, we can conclude that BERT behaves better when general knowledge of the target sentence is already acquired during pre-training. 

\section{Conclusion}
Memorization of pre-training examples plays a very important role in the performance of BERT. Indeed, our PreCog, which measures how much memorized pre-training knowledge cover target examples, is highly correlated with BERT's performance in inference. PreCog can then be also used as a measure of confidence for BERT-based decisions in downstream tasks.    

As BERT success is partially due to simple memorization of examples and given the overwhelming presence of ChatGPT, one area of future research should be on better understanding the relation between actual training examples and inferences in order to give credit to knowledge producers. 

%\newpage

\section*{Limitations}
This paper presents a small, focused contribution towards the understanding of the relation between memorization and performance of pre-trained language models (PTLMs). However, we leave some issues unresolved for this more long-term goal. Indeed, we have explored our idea only for a specific PTLM that is BERT with a specific pre-training task, that is, masked language model (MLM). Future analysis should explore whether our findings hold for other PTLMs based on MLM. Morever, we have not explored to what extent tasks examples are really covered by pre-training corpora used by PTLMs. The correlation between PreCog and the actual training examples should be investigated. Finally, PreCog is not suitable for PTLMs that are based pre-training tasks that ar not MLM. Then, other coverage measures should be defined in those cases.

%\section*{Ethics Statement}

%Accounting for the use of pre-existing knowledge by machine learning is a must. Indeed, modern solutions for downstream tasks have two main components: (1) a learning machine; (2) the knowledge that feeds the learning machine. In the current model, only the learning machine takes all the credit along with those who have contributed to define and customize the machine with learning data. Producers of learning data, which constitute previous knowledge, are not taken into consideration. The advent of ChatGPT, along with information retrieval systems like Bing, can make this problem even worse. Our study showed that part of the success of PTLMs is due to simple memorization of examples. Therefore, our work gives a basis to start the idea that knowledge producers should have credit in every inference of PTLMs and, more in general, of machine learning models. 

%Scientific work published at ACL 2023 must comply with the ACL Ethics Policy.\footnote{\url{https://www.aclweb.org/portal/content/acl-code-ethics}} We encourage all authors to include an explicit ethics statement on the broader impact of the work, or other ethical considerations after the conclusion but before the references. The ethics statement will not count toward the page limit (8 pages for long, 4 pages for short papers).

%Bibliography
%\nocite{*} % to test all bib entrys
\bibliographystyle{acl_natbib}  
\bibliography{custom,anthology} 

%\section{Example Appendix}
%\label{sec:appendix}

%This is a section in the appendix.

\end{document}